# Tree-based local explanations of machine learning model predictions – AraucanaXAI

E. Parimbelli[1][0000-0003-0679-828X], G. Nicora[1][0000-0001-7007-0862], S. Wilk[23][0000-0002-7807-454X], W. Michalowski[3][0000-0002-9198-6439], R. Bellazzi[1][0000-0002-6974-9808]

[1] University of Pavia, Italy
[2] Poznan University of Technology, Poznan, Poland
3 Telfer school of management, University of Ottawa, ON, Canada
enea.parimbelli@gmail.com

**Abstract.** Increasingly complex learning methods such as boosting, bagging and deep learning have made ML models more accurate, but harder to understand and interpret. A tradeoff between between performance and intelligibility is often to be faced, especially in high-stakes applications like medicine. In the present article we propose a novel methodological approach for generating explanations of the predictions of a generic ML model, given a specific instance for which the prediction has been made, that can tackle both classification and regression tasks. Advantages of the proposed XAI approach include improved fidelity to the original model, ability to deal with non-linear decision boundaries, and native support to both classification and regression problems.

## 1 Introduction

In recent years, an interesting trend in artificial intelligence (AI), now mostly known as eXplainable AI (XAI), has been gaining traction. Increasingly complex learning methods such as boosting, bagging and deep learning have made ML models more accurate, but harder to understand and interpret, often generating a contrast between performance and intelligibility (using a model which is a white-box, despite suboptimal performance)[1]. In order to tackle these problems, XAI investigates methods for analyzing or complementing AI models to make the internal logic and output of algorithms transparent and/or interpretable, making these processes and the predictions they generate humanly understandable and meaningful[1]. XAI is a re-emerging research trend, boosted by recently introduced regulations such as the European Union's GDPR[2], the US government's Algorithmic Accountability Act of 2019, or the U.S. Department of De-fense's Ethical Principles for Artificial Intelligence. Fairness, accountability, and transparency-related requirements are essential in AI applications to medicine (AIM), where AI often supports high-stakes decisions in diagnosis, prognosis and treatment[2], resulting in explanation facilities being an integral part of early AIM expert systems[3].

---

[1] https://www.ai4eu.eu/simple-guide-explainable-artificial-intelligence, (2021/04/09)
[2] General Data Protection Regulation

In this paper we present a novel methodological approach for local, model-agnostic explanations of predictive models, which is able to tackle both classification and regression tasks. We named our proposed approach AraucanaXAI since the generated explanations are based on Classification And Regression *Trees* (CART), with a reference to the *Araucaria Araucana* tree species, native to central and southern Chile, which is commonly known as the "monkey puzzle tree" in English-speaking countries.

## 1.1 Related work

Given the length restrictions, we limit our overview of related work to directly comparable approaches for model-agnostic, post-hoc local explanation approaches. We point readers interested in a more comprehensive overview of XAI methods to [4].

Among the most popular approaches for local interpretability, LIME [5] generates local explanations by fitting a linear model to the neighborhood of the test instance at hand $x$, weighting the neighboring samples according to their distance from $x$. In turn, LIME uses such linear local *explanator* model to present feature importance as an *explanation* to the user. LIME's strengths consist in its use of linear models which are inherently *white-box*, its being model-agnostic (i.e. applicable to any class of ML) and its polished implementation, which is widely available for use. Linear models are possibly even more suited for XAI in the medical community where concepts like odds ratios, and importance of linear independent predictors are widely understood.

Another notable XAI approach is SHAP[6] which used Shapley values, a concept drawn from game theory, for generating explanations of predictions. SHAP's explanations strive for being actionable and counterfactual, and are thus well-suited for medical applications, as demonstrated in [7]. [6] has also proposed a computationally efficient method to apply SHAP restricting it to tree-based ML models.

Lakkaraju and colleagues[8] proposed an XAI approach, later named BETA, based on decision sets to generate local as well as global model-agnostic explanations. The authors also define an experiment measuring quantifiable properties like fidelity and interpretability of the generated explanations. To the best of our knowledge, IDS nor BETA provide an open implementation, which makes comparisons harder.

## 2 Methods

Our AraucanaXAI approach is based on a relatively small set of principles: i) use trees in order to generate the *explainer* as a white-box model, with the ability to deal with non-linear decision boundaries, ii) grow an unpruned tree in order to maximise fidelity of the explanation iii) generate local explanations starting from a set of neighboring instances, coming from the original training set, re-labelled with the predictions of the *explained* model iv) use the *explainer* tree as the proposed explanation. Algorithm 1 presents the pseudocode of AraucanaXAI, which implement these principles.

---
**Algorithm 1** Local Tree-based explanation
---
**Require:**
    classifier $f$, number of local neighbors $N$, instance $x$, distance function $dist$

1: Compute $D = dist(x, z)$ for each training element $z$
2: Select the subset $T_n$ as the first $N$ training samples with lowest distance from $x$
3: Assign to each samples in $T_n$ the class predicted by $f$
4: Create explainer set $S$, augmenting $T_n$ with additional examples using SMOTE oversampling. Then classify $S$ using $f$ and use the predictions as new labels for the explainer set $S$
5: Train $e$ as a decision tree without pruning on $T_n \cup x \cup S$
6: Extract IF-THEN rules from $e$ and provide explanation for $x$ classification

---

## 2.1 Experimental setup

We developed a first POC of our approach on the widely used MIMIC dataset[9], which collects deidentified health data associated with thousands of intensive care unit (ICU) admissions. MIMIC includes clinical data, vital signs, laboratory tests, medications and other features that can be exploited by machine learning for in-hospital mortality prediction. After removing features with a high percentage (>=90%) of missing values and removing patients with at least a missing value in the remaining attributes, a total of 5248 patients with 48 features are included in our analysis. The binary dataset is highly imbalanced (14% of patients are reported in the "in-hospital death" class). We performed a 5-fold cross validation to train and evaluate a Gradient Boosting classifier, but our framework can be applied to any type of black box machine learning model. The mean accuracy reported with the cross validation is 82%, while the mean specificity and recall are respectively 86% and 60%.

However, in this case we are not interested in achieving the best possible performance of our classifier, but in the interpretation of individual predictions by the model. We therefore select one test fold, containing 1049 examples, and we applied the algorithm reported in Figure 1. Briefly, for each test instance *x*, we select the first *N* training samples closest to *x*, we classify the selected samples with the trained Gradient Boosting model and we apply the SMOTE algorithm to augment the local training set that will be used to fit an explainer decision tree without pruning. In our experiment, we set N=100. To handle the presence of categorical features, we select the *gower* function as our distance function and we apply the SMOTE-NC algorithm, specific for datasets containing numerical and categorical features (implementation available in the imblearn python package). Once the local dataset around the test instance *x* is augmented, a decision tree classifier is trained using CART (scikit-learn implementation) and IF-THEN rules for classification are extracted. We also evaluated the fidelity of the explainer model, testing whether its prediction on each *x* is consistent with the class predicted by the Gradient Boosting model. We compare the result of our fidelity test with the same experimental setup using LIME as method for generating the explanations.

The POC experiment is implemented in Python 3.7. Gradient Boosting and Decision Tree classifiers implementations are provided by scikit-learn package. The source code is available in a Colab notebook: https://bit.ly/3a7raQs.

## 3  Results

To qualitatively evaluate the results of the local tree-based explanation, we report the explanation generated by AraucanaXAI for a particular test patient whose class is 0 (= "no-death"), in the notebook accompanying the article. Then, we evaluate whether the classification performed by the explainable model are consistent with those made by the black box one, i.e. its fidelity. In fact, there is no structural guarantee that an *explainer* model will generate the same prediction of the *explained*, more complex and better performing model. Yet, in our experiment, all of the ~1000 test set classifications made by the AraucanaXAI explainer agree with the classification of the original Gradient Boosting model. On the other hand LIME, which uses linear explainer models, provides a consistent classification in ~90% of cases when tested on the same fidelity-testing task. Complete results and specific settings of the POC experiment are available in the Python notebook complementing the article: https://bit.ly/3a7raQs.

## 4  Discussion

Our AraucanaXAI approach has a number of advantages over comparable approaches for local, model-agnostic, post-hoc explanations. These include improved fidelity over LIME[5] and IDS/BETA[8], as demonstrated by the experiment reported in the present article and a previously published, independent experiment specifically targeting fidelity of XAI approaches[10]. Additionally, we provide a POC implementation of AraucanaXAI, in order to improve reproducibility and transparency the approach and experiment presented in this article. Also, thanks to CART being able to handle both classification and regression problems, our approach natively tackles both supervised learning class of problems. Finally, since AraucanaXAI actively re-uses the original training set of the explained model, it has a better chance to uncover possible biases and unexpected model behaviors which are due to the specifics not only of the learned model, but also of the dataset used for its training.

At the same time, we acknowledge the following limitations. Since we grow unpruned explainer trees, we might end up with large, complex explainer models that are difficult to read and interpret, even when converted in decision rules. Furthermore, our approach is limited to the explanations of single predictions, i.e. local explanations. Furthermore, our approach assumes the original training set is available along with the final learned model to be explained. Finally, our choice of employing SMOTE-NC to obtain a denser explainer set *S*, results in the introduction of artificial instances in *S*, which is subject to the known limitations of such a practice. Some of these limitations are structural, while others are planned to be addressed with future work. For example we may combine our approach with TreeExplainer[6], in order to improve the presentation of explanations based on large unpruned *explanator* trees. Currently, our priority is set on implementing AraucanaXAI as a distributable library, to facilitate its re-use and enabling adoption in further independent experiments. Secondly, we plan to validate the quality and human-friendliness of explanations, specifically with clinician-users, with a human-in-the-loop validation approach.

## 5  Conclusion

Local, model-agnostic post-hoc explanations constitute a valuable effort towards tackling the performance vs interpretability tradeoff. In this article we proposed a novel approach named AraucanaXAI, which allows generation of tree-based explanations for specific predictions while providing improved fidelity to the original model predictions.

**References**

1. Caruana, R., Lundberg, S., Ribeiro, M.T., Nori, H., Jenkins, S.: Intelligible and Explainable Machine Learning: Best Practices and Practical Challenges. In: P 26th ACM SIGKDD. pp. 3511–3512.ACM, New York, NY, USA (2020).
2. Holzinger, A., Langs, G., Denk, H., Zatloukal, K., Müller, H.: Causability and explainability of artificial intelligence in medicine. WIREs Data Mining and Knowledge Discovery. 9, e1312 (2019). https://doi.org/10.1002/widm.1312.
3. Shortliffe, E.H., Davis, R., Axline, S.G., Buchanan, B.G., Green, C.C., Cohen, S.N.: Computer-based consultations in clinical therapeutics: explanation and rule acquisition capabilities of the MYCIN system. Comput. Biomed. Res. 8, 303–320 (1975).
4. Guidotti, R., Monreale, A., Ruggieri, S., Turini, F., Giannotti, F., Pedreschi, D.: A Survey of Methods for Explaining Black Box Models. ACM Comput. Surv. 51, 93:1-93:42 (2018). https://doi.org/10.1145/3236009.
5. Ribeiro, M.T., Singh, S., Guestrin, C.: "Why Should I Trust You?": Explaining the Predictions of Any Classifier. arXiv:1602.04938 [cs, stat]. (2016).
6. Lundberg, S.M., Erion, G., Chen, H., DeGrave, A., Prutkin, J.M., Nair, B., Katz, R., Himmelfarb, J., Bansal, N., Lee, S.-I.: From local explanations to global understanding with explainable AI for trees. Nature Machine Intelligence. 2, 56–67 (2020). https://doi.org/10.1038/s42256-019-0138-9.
7. Lundberg, S.M., Nair, B., Vavilala, M.S., Horibe, M., Eisses, M.J., Adams, T., Liston, D.E., King-Wai Low, D., Newman, S.-F., Kim, J., Lee, S.-I.: Explainable machine-learning predictions for the prevention of hypoxaemia during surgery. Nat Biomed Eng. 2, 749–760 (2018). https://doi.org/10.1038/s41551-018-0304-0.
8. Lakkaraju, H., Bach, S.H., Jure, L.: Interpretable Decision Sets: A Joint Framework for Description and Prediction. KDD. 2016, 1675–1684 (2016). https://doi.org/10.1145/2939672.2939874.
9. Johnson, A.E., Pollard, T.J., Shen, L., Li-wei, H.L., Feng, M., Ghassemi, M., Moody, B., Szolovits, P., Celi, L.A., Mark, R.G.: MIMIC-III, a freely accessible critical care database. Scientific data. 3, 160035 (2016).
10. Lakkaraju, H., Kamar, E., Caruana, R., Leskovec, J.: Interpretable & Explorable Approximations of Black Box Models. arXiv:1707.01154 [cs]. (2017).